%% file: 000main.tex
\documentclass{bmvc2k}

\title{Evolving Image Compositions for Feature Representation Learning}

\addauthor{Paola Cascante-Bonilla}{pc9za@virginia.edu}{1}
\addauthor{Arshdeep Sekhon}{as5cu@virginia.edu}{1}
\addauthor{Yanjun Qi}{yq2h@virginia.edu}{1}
\addauthor{Vicente Ordonez}{vicenteor@rice.edu}{2}

\addinstitution{
 University of Virginia\\
 Charlottesville, Virginia, USA
}
\addinstitution{
 Rice University\\
 Houston, Texas, USA
}

\runninghead{Cascante-Bonilla et al.}{Evolving Image Compositions}

\usepackage{times}
\usepackage{epsfig}
\usepackage{graphicx}
\usepackage{amsmath}
\usepackage{amssymb}
\usepackage{paralist}
\newcommand{\R}{\mathbb{R}}

\usepackage{pifont}
\newcommand{\xmark}{\text{\ding{55}}}
\usepackage{booktabs}

\usepackage{array,multirow}
\newcommand{\STAB}[1]{\begin{tabular}{@{}c@{}}#1\end{tabular}}

\usepackage{algorithm}
\usepackage{algorithmicx}
\usepackage[noend]{algpseudocode}
\algdef{SE}[DOWHILE]{Do}{doWhile}{\algorithmicdo}[1]{\algorithmicwhile\ #1}

\input{preambular}

\begin{document}

\maketitle


\input{000abstract}

\input{010introduction}

\section{Related Work}
\label{sec:background}
\input{020background}

\section{Our Method}
\label{sec:meth}
\input{030method}

\input{040experiment}

\input{042implementation}

\input{043ImageClassify}
\input{044localization}
\input{045transfer}

\input{046robustness}
\input{052ablationAnalysis}

\input{053ablationFitnessFunc}

\vspace{-0.2in}
\section{Conclusion}
\label{sec:concl}
\input{070conclusion}

\section{Acknowledgements}
The authors would like to thank the anonymous reviewers for their insightful comments. This work is supported by the National Science Foundation under awards No \#2045773 and \#2040961.

\newpage

\bibliography{egbib}

\newpage
\appendix
\input{080appendix}
\input{051searchAnalysis}

\input{081appendix}

\end{document}

%% file: preambular.tex
\usepackage{xcolor}

\usepackage{flushend}

\def\x{{\mathbf x}} 
 
\def\y{{\mathbf y}}

\usepackage[utf8]{inputenc}

%% file: 000abstract.tex
\begin{abstract}

Convolutional neural networks for visual recognition require large amounts of training samples and usually benefit from data augmentation.
This paper proposes PatchMix, a data augmentation method that creates new samples by composing patches from pairs of images in a grid-like pattern. These new samples are assigned label scores that are proportional to the number of patches borrowed from each image. We then add a set of additional losses at the patch-level to regularize and to encourage good representations at both the patch and image levels. A ResNet-50 model trained on ImageNet using PatchMix exhibits superior transfer learning capabilities across a wide array of benchmarks. Although PatchMix can rely on random pairings and random grid-like patterns for mixing, we explore evolutionary search as a guiding strategy to jointly discover optimal grid-like patterns and image pairings. For this purpose, we conceive a fitness function that bypasses the need to re-train a model to evaluate each possible choice.  In this way, PatchMix outperforms a base model on CIFAR-10 (+1.91), CIFAR-100 (+5.31), Tiny Imagenet (+3.52), and ImageNet (+1.16).

\end{abstract}

%% file: 010introduction.tex
\section{Introduction}
\label{sec:intro}

\begin{figure}[t] 
\centering
\includegraphics[width=0.92\textwidth]{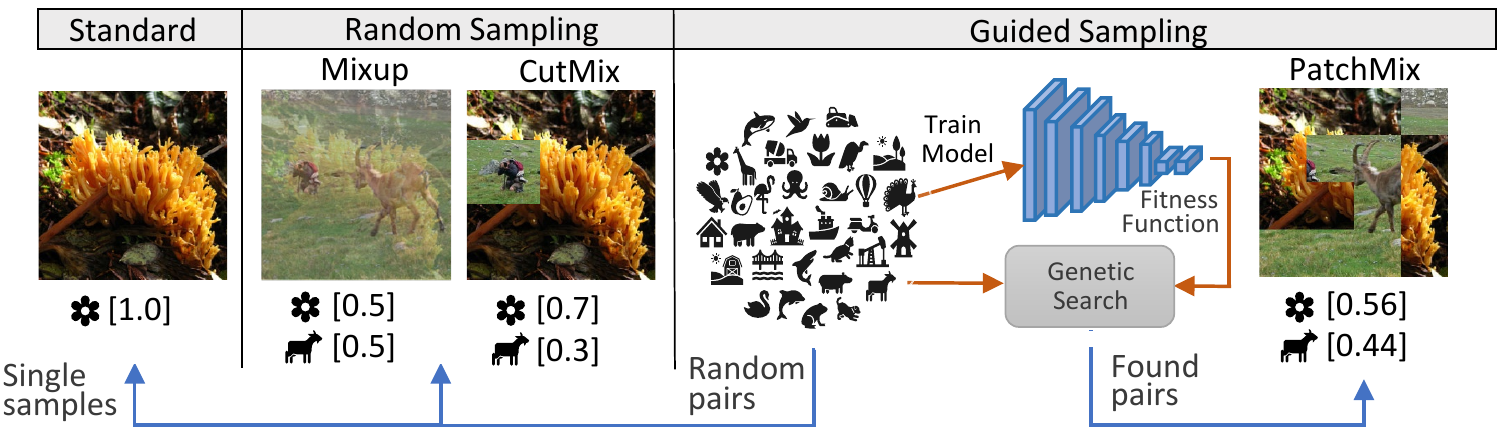}
\caption {Examples of different data augmentation techniques. Our proposed method exploits patch-based image compositions that allows flexible combinations for data augmentation. PatchMix allows training a model that can be used as a fitness function of an evolutionary search pipeline to find optimum mask configurations and sample pairs. The numbers below each image correspond to the new labels, which are associated  with the proportion in which each class has been mixed. 
\label{fig:illustrative_aug_methods}
\vspace{-0.2in}
}
\end{figure}

Deep convolutional neural networks (CNNs) have pushed forward significant progress in many computer vision tasks \cite{Russakovsky2015ImageNetLS, Krizhevsky2017ImageNetCW, He2016DeepRL, Ren2015FasterRT, Wang2020RDSNetAN}.  These high-capacity models tend to memorise  their training data to some extent, therefore, they might lead to suboptimal generalization.  Recent work has proposed various data augmentation techniques to alleviate this issue by smoothing out the input space, the output space, or both.
Relevant literature falls roughly into two groups: (1) Data augmentation from individual input samples~e.g.~\cite{Zoph2017NeuralAS, Lu2019NSGANetNA, Pham2018EfficientNA}, and (2) Data augmentation that creates new samples by interpolating pairs of samples~e.g.~\cite{Zhang2018mixupBE,Mai2019MetaMixUpLA,Faramarzi2020PatchUpAR}. Our paper focuses on the second line of work and proposes to interpolate two samples via  patch-level compositions in a grid pattern. Figure~\ref{fig:illustrative_aug_methods} shows examples of using data augmentation strategies to create samples.

Multiple successful strategies have been proposed for combining pairs of samples to regularize deep learning models such as Mixup~\cite{Zhang2018mixupBE}, Cutmix~\cite{Yun2019CutMixRS}, and Cutout~\cite{Devries2017ImprovedRO, chen2020gridmask, singh2018hideandseek}. These studies have shown patch-level perturbations and augmentation strategies to be beneficial for robust feature representation learning in vision. Our method, we call PatchMix, provides two novel designs when compared to the recent works. First, PatchMix uses a grid mask to decompose the image space into a regular grid of patches. Image compositions via a grid mask allows for greater diversity in making the augmented samples (as shown in Figure~\ref{fig:illustrative_aug_methods}). We also add an auxiliary patch-level supervision on top of the image-level supervision to encourage better and more robust representation learning (refer to  Figure~\ref{fig:fig_gridmix}).  

Second, different from the previous methods relying on heuristics to find patches to combine~\cite{GridMix2021}, we introduce a guided search strategy via genetic search to find the best set of category pairs to mix and to search for optimal category-dependent grid masks for combining pairs of images. 
Our search aims to find a set of category pairs (and corresponding masks) for interpolating pairs of image samples to create new samples, hoping to achieve improved model training and generalization. One main challenge when using genetic search for pairwise sample interpolation is the expensive computation cost. This is because evaluating the fitness of each interpolation configuration requires training and evaluating a new model. We, instead, propose a computationally feasible approximation to calculate such a fitness calculation,  avoiding the bottleneck of retraining the model for each potential configuration.  While genetic search has been used for exploring single sample data augmentation, to the best of our knowledge, our work is the first to explore evolutionary techniques to find the best configurations over the space of interpolations between samples.

Empirically, we validate the effectiveness of PatchMix on the regular image classification task (via CIFAR~\cite{cifar10, cifar100}, Tiny Imagenet~\cite{tiny_imagenet_Le2015TinyIV} and ImageNet~\cite{imagenet_cvpr09}), on the weakly-supervised localization (WSOL) task (via CUB-200-2011~\cite{Wah2011TheCB}), the object detection task (via Pascal VOC~\cite{VOC}), on the transfer learning task (via CUB-200-2011, SUN397~\cite{SUN} and multi-label datasets: Pascal VOC and MS-COCO~\cite{COCO} and NUS~\cite{NUS}),
and on the image captioning task (via MS-COCO). Finally, we show consistent robustness results on a model trained with PatchMix when tested against adversarial examples using the Fast Gradient Sign Method (FGSM)~\cite{whiteBoxAttack43405} white box attack.

%% file: 020background.tex

\subsection{Pairwise data augmentation}
Mixup~\cite{Zhang2018mixupBE} was the first work that proposed the idea of interpolating two images, and their ground truth labels to augment the training data. Mixup and its variants may suffer from the issue of object local ambiguity, also called \textit{manifold intrusion}~\cite{Guo2019MixUpAL}. This occurs when the objects inside two image samples are interpolated in such a way that introduce visual confusion, and the true labels contradict the synthetic labels of the generated mixed sample. However, this method has proved effective and general, almost always providing some improvement over a baseline that relies only on single sample data augmentation strategies.

Recent literature has tried a set of mechanisms to deal with the \textit{manifold intrusion} problem \cite{Guo2019MixUpAL} by proposing different data interpolation alternatives to Mixup. 
For example, ManifoldMixup \cite{Verma2019ManifoldMB} and PatchUp \cite{Faramarzi2020PatchUpAR} interpolate the hidden states instead of the input space. MetaMixUp \cite{Mai2019MetaMixUpLA} proposes to use meta-learning to learn a mixing coefficient that could avoid a high frequency of cases of manifold intrusion. \cite{Dabouei2020SuperMixST} train an extra neural network to anticipate whether a particular combination of two images may suppress information or add manifold intrusions. \cite{Chou2020RemixRM} propose to force a balanced sampling from the training set for selecting the images to be interpolated. CutMix and variants create random binary masks to sample a patch and to apply the corresponding image interpolation \cite{Yun2019CutMixRS, Harris2020FMixEM} only on a subregion. Our paper proposes a new strategy, PatchMix, that allows for sampling multiple patches from an image to interpolate with a second image. Moreover, we select optimal interpolations that are on average more challenging than random interpolations using genetic search.

More recently, other methods~\cite{Walawalkar2020AttentiveCA, Faramarzi2020PatchUpAR} such as GridMix~\cite{GridMix2021}, have explored patch-like masks to enable input samples interpolations along with their corresponding labels. 
In our proposed method, we take the last layer of the CNN and divide it as a matrix where each patch also corresponds to the Patch-Mask we use to mix the input samples. In our case, our patch-loss is equally weighted into the whole pipeline and we try to solve the \textit{manifold intrusion} problem during training. This issue happens when synthetic samples generated from interpolating two real samples are assigned a label that contradicts the individual samples. Additional analysis about the manifold intrusion could be found in the Appendix, showing its effect in the decision boundary for a three-way classifier on synthetic data when using different interpolations such as Mixup, CutMix, and our proposed PatchMix. 

\subsection{Samplewise data augmentation}
Another popular group of data augmentation research explores ways to augment samples via individual one-to-one sample transformations. Multiple recent works apply random transformations over an image to augment training data. These transformations range from random cropping, flipping, or rotating an image~\cite{Takahashi2018DataAU}, to random erasing~\cite{Devries2017ImprovedRO, Zhong2020RandomED}, and even more complex random transformations~\cite{Xie2019UnsupervisedDA}. More recently, researchers have proposed methods to automatically search for data augmentation policies with Reinforcement Learning (RL) or Evolutionary Algorithms~\cite{Zoph2017NeuralAS, Lu2019NSGANetNA, Pham2018EfficientNA}. This idea also relates to using RL systems to find state-of-the-art model architectures for image classification~\cite{Zoph2018LearningTA} using policy gradient optimization methods~\cite{Schulman2017ProximalPO}. 
This setup is typically expensive due to the need to retrain the model for evaluating all sub-policies or configurations~\cite{Cubuk2019AutoAugmentLA, Lim2019FastA}. Differently, PatchMix generates augmented images from pairs of samples and uses genetic search that is guided by a novel fitness criteria based on the difficulty of the chosen configurations. 


%% file: 030method.tex

PatchMix includes three components:  (1) A patch mask that enables a grid-level composition involving all patches from two images. We decompose the image space into a grid of regular-sized patches and design a binary mask on each patch position controlling the composition. (2) A new loss function that enforces patch-level label supervision, in addition to the global image supervision. This loss enhances the regularization provided by our patch-based sample augmentation and provides a useful fitness function for evaluating what candidate patches to use.  (3) A search strategy based on genetic search to find the best patch mask for combining pairs of image categories.

\begin{figure*}[h] 
\centering
  \includegraphics[width=.92\textwidth]{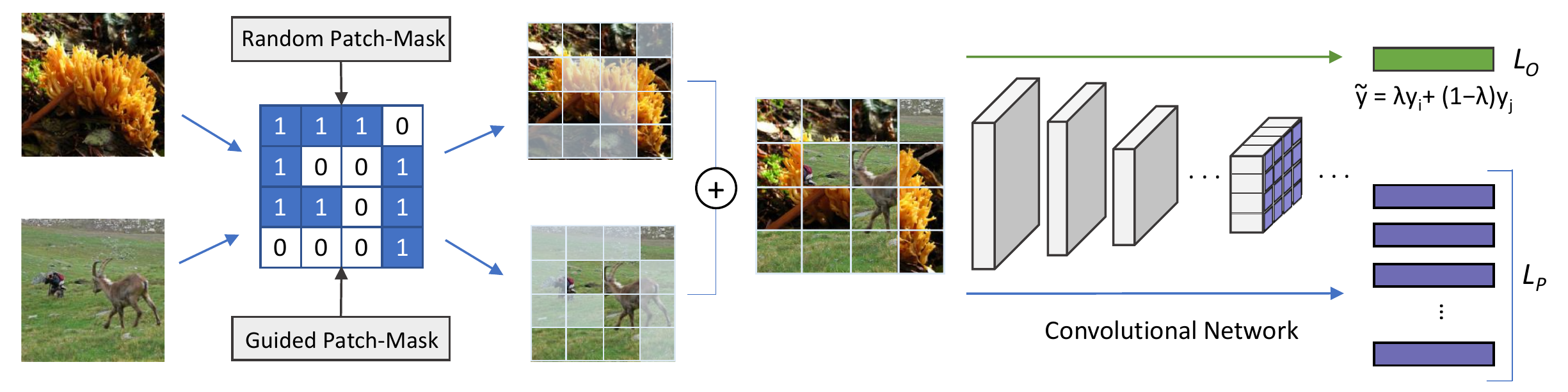}
\caption {Overview of PatchMix. We first create a binary mask \textit{M} with \textit{P}x\textit{P} number of patches using our proposed PatchMix strategies: Guided PatchMix and its variation Random PatchMix (see Section \ref{sec:meth} for details on how to obtain the mask). This mask is then used to interpolate two random images which will create a new image sample. This generated sample is used to train the model, a CNN with the last convolutional layer modified to create \textit{P}x\textit{P} patches of equal size. In this way, we are able to output the values corresponding to the input patches corresponding to each image, and the mixed output of the whole new image. }
\label{fig:fig_gridmix}
\vspace{-0.2in}
\end{figure*}

\subsection{Binary Patch-Mask $\mathbf{M}$}

Let $\x \in \R^{W \times H \times C}$ denote a training image  and $\y$ be its corresponding label. The goal of PatchMix is to synthesize additional training samples by interpolating  pairs of inputs. For example, for samples $\x_i$, $\x_j$ and their corresponding labels $\y_i$, $\y_j$, we use a patch mask matrix $\mathbf{M} \in \{0,1\}^{W \times H}$ to create a new sample $(\Tilde{\x}, \Tilde{\y})$:
\begin{align}
    \Tilde{\x} & =  \mathbf{M} \odot \x_{i} + (1- \mathbf{M}) \odot \x_{j}, \\
    \Tilde{\y} & =  \lambda \y_i + (1-\lambda) \y_j,
\label{eq:gridmix}
\end{align}
where $\lambda = \sum_{s=1}^W \sum_{t=1}^H {\mathbf{M}(s,t)}/(W\times H)$. More specifically, we divide the mask $\mathbf{M}$ into $P^2$ patches, resulting in each patch region of size $W/P \times H/P$. We additionally constrain the values ${\mathbf{M}(s,t)}$ in each patch region to be the same values. In this way, we force the two input images to be interpolated using patches of the same size, with $2^{P \times P}$ possible configurations. This process is illustrated in Figure~\ref{fig:fig_gridmix} (on the left side) for a given Patch-Mask with $P=4$. 

\subsection{Patch-level Supervision}

The key intuition of PatchMix is that image patches provide strong supervisory signals. 
We, therefore, design mechanisms to exploit such weak supervisions for each of the image patches.  We design a Convolutional Neural Network, that takes as input $\Tilde{x}$, so that its last convolutional layer produces a set of feature region vectors corresponding to each of the input $P^2$ patch regions. In this way, we force the network to output $P^2$ additional class predictions corresponding to labels for each individual patch. Thus, we adopt the following objectives for each generated sample:
\begin{align}
L_O & =  -\sum_{i=1}^{C} \Tilde{y}_i \log(\hat{y}_i), \\
    L_P & =  -\sum_{n=1}^{P^2} \sum_{i=1}^{C} \Tilde{y_i}_{n} \log(\hat{y_i}_{n}), \label{eq:lp} \\
    L_T & = (L_O + (L_P / P^2)) / 2, 
\label{eq:loss}
\end{align}
where $C$ is the number of classes, $L_O$ corresponds to the cross entropy loss over the image-level label vector $\Tilde{\y}$, $L_P$ corresponds to a sum of the cross entropy losses for each patch with respect to patch-level (pseudo) labels. Here we assume for the $n$-th patch, its label  $\Tilde{\y}_n$ is the same as its image label. $L_T$ is the combined loss using both image-level and patch-level supervision. 
Figure \ref{fig:fig_gridmix} shows the full process of combining two images given a fixed patch interpolation mask for a pair of input images.

\subsection{Evolutionary Search over Interpolations}

In order to train a model well under PatchMix, we need to define the patch masks $\mathbf{M}$ for combining a pair of samples. In its most basic form we can train a model by selecting random pairs of images from two arbitrary categories under a mask $\mathbf{M}$ such that the entries for each patch region are sampled from a beta distribution $B(\alpha, \alpha)$. We refer to this basic formulation as \textbf{Random PatchMix}. Further, we propose a better strategy to search for the optimal masks $\mathbf{M}$ that help us interpolate pairs of samples optimally. We refer to this as \textbf{Guided PatchMix}. In Guided PatchMix, we search for a specific mask $\mathbf{M}_{ij}$ for a pair of image categories $(c_i, c_j)$. We propose a novel genetic search optimization to automatically identify a set of category pairs $(c_i, c_j)$ that are good to interpolate, and the set of mask $\mathbf{M}_{\cdot,\cdot}$ that determine how their images mix to generate new samples. 


We have two concrete search goals in Guided PatchMix. 
\begin{compactitem}
\vspace{0.02in}
\item (a) To identify what pairs of image categories are suitable for mixing.
\item (b) For a specific category pair $(c_i, c_j)$, what is the best mask $\mathbf{M}_{\cdot,\cdot}$ that allows for their images to interpolate well so that they generate new samples resulting in some improved generalization.
\vspace{0.02in}
\end{compactitem}

We, therefore, represent an \textbf{individual} candidate solution in our search as follows: (a) It includes a set of active class combinations $A = \{(c_i, c_j)\}$ and $|A|<=N$ (here $N$ is a hyperparameter to tune). (b) For each active class pair $(c_{i_a}, c_{j_a})$ in the active set $A$, we have a mask matrix $\mathbf{M}_a$  of dimension $P \times P$ to search for, presenting $2^{P \times P}$ possible configurations to combine images from class $(c_{i_a}$ and  $c_{j_a})$. Our \textbf{population} is initialized with \textit{I} different individuals. Each individual $A_i$ is built from random pairing between classes and from assigning random binary values in each $\mathbf{M}_a$. These individuals are evolved for approximately \textit{G} \textbf{generations}. We decided to limit the amount of active combinations to size $N$, in order to narrow down the growth of the search space; this decision helps the algorithm to converge faster and yield better compositions. We show one cycle of our genetic search implementation in Figure \ref{fig:fig_genetic_search} in the Supplementary Material.

One key component in using an evolutionary framework is to require a cost-effective \textbf{fitness function} for evaluating each ``individual" candidate solution. For our search, we can evaluate whether a specific ``individual" (i.e., a set of active class combinations and their interpolation masks) is effective or not by training a model that uses those for data augmentation. The resulting accuracy of such a model serves as a good fitness criterion. However, it is computationally too expensive to train models for each possible ``individual" configuration, considering the vast search space for possible set $A = \{(c_i, c_j)\}$ with size $N$ plus the vast configuration space to search for each mask $\mathbf{M}_{c_{i_a}, c_{j_a}}$. Instead, our fitness criteria uses the average of the patch-level scores (based on Equation~\ref{eq:lp}) from $f_T$ on the validation set to choose ``individual" solutions that are challenging and thus yielding potentially more informative interpolations than random patch selections.



After evaluating the fitness scores on each individual in the current population, some of them are discarded. Then some pairs of individuals are combined using a \textbf{crossover} function. Our choice of crossover function combines corresponding masks $\mathbf{M}_{ij}$ from two different surviving individuals by copying the left and right half of each mask. 
Furthermore, some of these new \textbf{offspring} are transformed using a \textbf{mutation} operation with a low random probability. We define a set of possible mutation operations. 
Figure \ref{fig:fig_mutations} in the Supplementary Materials shows examples of mutation operations. The search algorithm stops after a specific number of generations, or early stops if there is no further improvement. This last condition typically happens when the offspring in the current generation are almost the same as in the previous generation.

\vspace{-0.1in}

\subsection{PatchMix Training Workflow}

In summary, guided PatchMix trains a model as follows: 
\begin{compactitem}
\item First phase: We train Random PatchMix to define our fitness criteria $f_T$ by optimizing $L_T$ over a dataset of images and their corresponding labels. 
\item Second phase: We use genetic search to find the best set of masks $\mathbf{M}_{i,j}$ and active category pairs $(c_i, c_j)$ that correspond to each of the discovered class combinations by using the fitness criteria induced by $f_T$. 
\item Third phase: We use the best set of masks $\mathbf{M}_{i,j}$ learned by our evolutionary algorithm to create informative augmented training samples based on the class combinations $(c_i,c_j)$ discovered in the second phase; 
\item Fourth phase: We train a final prediction network $f_O$ using the original training set, a randomly-augmented set based on random masks similar to the one described in the first phase, and the augmented set sampled in the third phase. We train this function by minimizing the sum of losses $L_O$ over these samples. 
\end{compactitem}

%% file: 040experiment.tex
\vspace{-0.1in}

\section{Experimental Setup}
\label{sec:exp}


%% file: 042implementation.tex
\subsection{Implementation Details} \label{sec:settings}

\textbf{Evolutionary Search} 
We adopt DEAP~\cite{DEAP_JMLR2012}, an evolutionary computational framework, to work as our base genetic search data structure.  This framework allowed us to define each individual in the population as a set of vectors, along with their grid mask configurations. The population is set to 500 individuals, which are evolved for 250 generations. 
Each individual has a limited number of active combinations, we treat the total number of allowed active combinations ($N$) as a hyperparameter. In our experiments, we set this as equal to the number of classes in the dataset, along with the same class combinations $(c_i, c_i)$ pairs that are forced to be always active.
Since we set $P=4$ in all our experiments, each combination has $65,536$ possible configurations. We also set the crossover probability to 50\% and the mutation probability to 30\%. Since our fitness function is a model trained using Random PatchMix, we spawn 20 processes to work in parallel, each containing the trained network to evaluate each individual. These 20 processes ran on 5 servers, each with 4 NVIDIA GPUs (ranging from GTX1080, GTX1080 Ti and Titan X).

\textbf{PatchMix Training} We train for $400$ epochs, using mini-batches of $100$ images. All the networks are optimized using Stochastic Gradient Descent (SGD) with Nesterov momentum. We use a weight decay regularization of $0.0005$, a momentum factor of $0.9$, and an initial learning rate of $0.1$ which is updated using cosine annealing~\cite{cosineAnnealing}. In all our experiments, we set $P = 4$ and $\alpha = 1$.  

\textbf{Baselines} For both Mixup~\cite{Zhang2018mixupBE} and Cutmix~\cite{Yun2019CutMixRS}, we use $\alpha = 1.0$. A cropping region of $16 \times 16$ is used for Cutmix which is sampled from a Gaussian distribution with mean at the image centre. We also train for $400$ epochs, using mini-batches of $100$ images, SGD with nesterov momentum, a weight decay regularization of $0.0005$, a momentum factor of $0.9$, and an initial learning rate of $0.1$ which is updated using cosine annealing.

%% file: 043ImageClassify.tex
\vspace{-0.1in}

\section{Experimental Results}
\label{sec:results}

\subsection{Supervised Image Classification} \label{sec:cifar}
We evaluate PatchMix using both the random patch selections and our guided sampling strategy found using genetic search. Table~\ref{tab:class_results1} shows the top-1 accuracy and comparison against Mixup, Manifold Mixup and Cutmix, which are now standard techniques for data-augmentation and regularization on CIFAR-10 and CIFAR-100. Table~\ref{tab:class_results2} shows the top-1 accuracy and comparison against Mixup and Cutmix on Tiny-Imagenet and ImageNet. Random PatchMix outperforms a model trained without any data augmentation in all scenarios and is comparable to Mixup and Cutmix. Guided PatchMix outperforms all models trained using the other regularization approaches.

\begin{table}[h]
\setlength{\tabcolsep}{6.4pt}
\begin{center}
\scalebox{0.75}{
\begin{tabular}{l|c|c|c|c|c|c}
\hline
\multicolumn{7}{c}{CIFAR-10} \\
\hline
Model & Base & Mixup & Manifold & Cutmix & Rand      & Guided \\
       &      &       &  Mixup   &        & PatchMix   & PatchMix\\
\hline
 MobileNetV2 & $90.55 \pm 0.04$ & $91.39 \pm 0.02 $ & $91.79 \pm 0.11$ & $91.93 \pm 0.04$ & $92.64 \pm 0.02$ & $\textbf{93.85} \pm \textbf{0.07}$ \\
 ResNet32 & $92.61 \pm 0.03$ & $93.40 \pm 0.02 $ & $94.14 \pm 0.05$ & $93.92 \pm 0.06$ & $94.13 \pm 0.07$ & $\textbf{94.93} \pm \textbf{0.03}$ \\
 ResNet50 & $93.70 \pm 0.06$ & $94.75 \pm 0.03 $ & $95.24 \pm 0.06$ & $94.89 \pm 0.05$ & $95.04 \pm 0.08$ & $\textbf{95.48} \pm \textbf{0.02}$ \\
 ResNet56* & $93.95 \pm 0.04$ & $94.42 \pm 0.06 $ & $94.15 \pm 0.03$ & $93.92 \pm 0.07$ & $94.62 \pm 0.09$ & $\textbf{94.80} \pm \textbf{0.07}$ \\
 ResNet164* & $94.06 \pm 0.07$ & $95.12 \pm 0.03 $ & $95.55 \pm 0.08 $ & $95.72 \pm 0.07$ & $95.81 \pm 0.12$ & $\textbf{96.06} \pm \textbf{0.04}$ \\
\hline
\hline
 \multicolumn{7}{c}{CIFAR-100} \\
\hline
 MobileNetV2 & $66.55 \pm 0.21$ & $68.45 \pm 0.38$ & $68.97 \pm 0.41 $ & $69.14 \pm 0.39$ & $69.18 \pm 0.38$ & $\textbf{70.05} \pm \textbf{0.37}$\\
 ResNet32 & $68.52 \pm 0.38$ & $69.12 \pm 0.31$ & $70.82 \pm 0.39 $ & $71.32 \pm 0.30$ & $71.09 \pm 0.49$ & $\textbf{72.83} \pm \textbf{0.26}$\\
 ResNet50 & $71.37 \pm 0.27$ & $71.99 \pm 0.31$ & $72.65 \pm 0.40 $ & $72.91 \pm 0.36$ & $73.02 \pm 0.48$ & $\textbf{73.63} \pm \textbf{0.31}$\\
 ResNet56* & $71.60 \pm 0.37$ & $72.43 \pm 0.29$ & $73.21 \pm 0.47 $ & $74.02 \pm 0.35$ & $74.56 \pm 0.32$ & $\textbf{75.26} \pm \textbf{0.38}$\\
 ResNet164* & $72.43 \pm 0.22$ & $74.14 \pm 0.34$ & $75.07 \pm 0.49 $ & $76.97 \pm 0.31$ & $76.39 \pm 0.44$ & $\textbf{78.16} \pm \textbf{0.47}$\\
\hline
\end{tabular}
}
\end{center}
\caption{Results on supervised classification datasets. Base refers to each model trained without any interpolation technique. The asterisk (*) refers to PreAct-ResNet. All experiments were run 3 times, we report their mean and standard deviation.} 
\label{tab:class_results1}
\end{table}

\begin{table}[h]
\setlength{\tabcolsep}{4.6pt}
\begin{center}
\scalebox{0.75}{
\begin{tabular}{l|c|c|c|c|c|c|c|c|c|c}
\hline

\multicolumn{6}{c}{Tiny-Imagenet} & \multicolumn{5}{|c}{ImageNet} \\
\hline
Model & Base & Mixup & Cutmix & Random & Guided & Base & Mixup & Cutmix & Random & Guided\\
      &      &       &       & PatchMix & PatchMix      &      &       &       & PatchMix & PatchMix \\
\hline
ResNet50 & 61.18 & 63.04 & 63.36 & 62.94 & \textbf{64.70} & 76.27 & 77.01 & 77.41 & 77.38 & \textbf{77.43} \\
\hline
\hline
\end{tabular}
}
\end{center}
\caption{Results on supervised classification on ImageNet and Tiny-Imagenet. Base refers to each model trained without any interpolation technique.
} 
\label{tab:class_results2}
\end{table}

%% file: 044localization.tex
\vspace{-0.2in}

\subsection{Weakly Supervised Object Localization and Object Detection}

We also evaluate PatchMix on the weakly supervised localization task, which aims to find a target object using only the image-level label as supervision. In particular, we use the Class Activation Mapping (CAM)~\cite{choe2020evaluating} to extract the attention maps, and then we compute the maximal box accuracy, which is the bounding box accuracy and the Intersection over Union (IoU) of the proposed boxes, following the WSOL framework and evaluation benchmark recently proposed in~\cite{choe2020cvpr} referred to as MaxBoxAccV2~\footnote{https://github.com/clovaai/wsolevaluation}. 

PatchMix excels in this setting due to the patch-level supervision, which seems to give additional cues to the be considered by the scoring function. We show results on the CUB-200-2011 dataset trained on ResNet-50, VGG-16 and Inceptionv3 backbones. Random PatchMix outperforms other interpolation techniques as well as the baseline CAM. We show our results on Table~\ref{tab:wsol_detection_results}, and qualitative results in the supplementary materials (section~\ref{sec:qualitative_WSOL}). 
\begin{table}[h]
\setlength{\tabcolsep}{4pt}
\begin{center}
\scalebox{0.75}{
\begin{tabular}{l|c|c|c}
\hline
Method & ResNet50 & VGG16 & Inceptionv3  \\ 
\hline
Baseline (CAM) \cite{cvpr2016_zhou} & 63.0 & 55.6 & 56.7 \\
Mixup & 55.8 & 51.9 & 53.4 \\
Cutmix & 62.8 & 54.9 & 57.4 \\
\hline
PatchMix & {\bf 63.9} & {\bf 57.3} & {\bf 57.7} \\ 
\hline
\end{tabular}
}
\end{center}
\caption{ Results for the Weakly Supervised Object Localization task on the CUB-200-2011 dataset using three different backbones. The baseline is using the class activation mapping (CAM) without any data augmentation. We then apply Mixup, Cutmix and PatchMix, and report the MaxBoxAccV2~\cite{choe2020cvpr}. 
}
\label{tab:wsol_detection_results}
\end{table}

%% file: 045transfer.tex
\vspace{-0.2in}

\subsection{Transfer Learning Capacity} \label{sec:imagenet}

Table~\ref{tab:class_imagenet_results} presents results for various models pretrained on the ImageNet ILSVRC dataset and finetuned on seven different downstream tasks, including food recognition (Food-101\cite{food101_bossard14}), bird classification (CUB-200-2011~\cite{Wah2011TheCB}), scene recognition (SUN397~\cite{SUN}), multi-label object classification~(Pascal VOC~\cite{VOC}, COCO~\cite{COCO}, and NUS~\cite{NUS}), and image captioning (COCO Captions~\cite{chen2015microsoft}).  Our results include the performance for a base ResNet-50 model, a ResNet-50 model trained with CutMix and a ResNet-50 model trained with Random PatchMix. PatchMix shows the largest transferability across these tasks with the best performance scores in $7$ out of $8$ tasks. 

\begin{table*}[h]
\setlength{\tabcolsep}{3pt}
\begin{center}
\scalebox{0.75}{
\begin{tabular}{l|c|c|c||c|c|c||c|c}
\hline
 ImageNet    & Food-101 & CUB-200 & SUN397 & VOC & COCO & NUS & \multicolumn{2}{c}{MS-COCO NIC~\cite{Vinyals2015ShowAT}} \\
 Pretrained  & top-1 acc & top-1 acc & top-1 acc & mAP & mAP & mAP & BLEU-1 & BLEU-4 \\ 
\hline
RN  & 87.70 & 76.30 & 60.41 & 92.13 & 79.64 & 80.19 & 61.4 & 22.9 \\
RN+M  & 87.82 & 78.91 & 60.16 & 91.80 & 81.20 & 81.72 & 61.6 & 23.2 \\
RN+CM  & {\bf 88.02} & 77.77 & 58.48 & 92.41 & 79.68 & 80.53 & 64.8 & 24.9 \\
RN+PM & 87.95 & 79.17 & {\bf 61.08} & {\bf 92.42} & 81.27 & 82.45 & {\bf 66.8} & {\bf 26.3} \\
RN+GPM & 87.50 & {\bf 79.50} & 60.99 & 92.39 & {\bf 83.21} & {\bf 82.95} & 65.5 & 25.5 \\
\hline
\end{tabular}
}
\end{center}
\caption{ Transfer learning results on different datasets. We use a ResNet-50 model pretrained on ImageNet via four different training strategies. The first row corresponds to normal training without data-pair interpolations. RN+M, RN+CM, RN+PM and RN+GPM refers to finetuning a ResNet-50 model with Mixup, CutMix, PatchMix and Guided PatchMix respectively.  } 
\label{tab:class_imagenet_results}
\end{table*}


%% file: 046robustness.tex
\subsection{Robustness} \label{sec:robustness}

We perform studies on the FGSM~\cite{whiteBoxAttack43405} white box attack on ImageNet using $\varepsilon$ = 0.1, 0.2, 0.3. The aim of this test is to create adversarial samples by fixing the perturbation on a pixel to be of a fixed size (i.e. $\varepsilon$). As shown in Table~\ref{tab:robustness_results}, PatchMix consistently outperforms previous methods in 2 out of 3 attacks.

\begin{table}[!th]
\setlength{\tabcolsep}{2.4pt}
\begin{center}
\scalebox{0.65}{
\begin{tabular}{c|c|c|c|c|c}
\hline
$\varepsilon$ & Base & Mixup & Cutmix & Random & Guided \\
              &      &       &       & PatchMix & PatchMix \\
\hline
0.1 & 15.96 & 28.42 & 29.26 & 30.62 & \textbf{31.88} \\
0.2 & 9.12 & 20.45 & 19.92 & 21.07 & \textbf{21.68} \\
0.3 & 5.87 & \textbf{15.31} & 13.65 & 14.29 & 14.57 \\
\hline
\end{tabular}
}
\end{center}
\caption{Results on the FGSM white box attach on ImageNet: we report the top-1 accuracy a ResNet-50 model trained with ImageNet using all techniques. The $\varepsilon$ indicates the perturbation level of the adversarial images generated.
} 
\label{tab:robustness_results}
\end{table}

%% file: 052ablationAnalysis.tex
\subsection{Ablation Studies} 
\label{sec:ablation_studies}

Given the flexible capabilities our grid-mask design allows, we conduct a thorough set of experiments to determine whether the patch-level loss is helpful or not, and to what extent the number of patches impact the overall performance.
To examine all of these possibilities, we run a set of experiments on CIFAR-10 using ResNet-32 as the base network. For these experiments we keep the same hyperparameter selections we use to report our results in section~\ref{sec:results}. 
We vary the size of $P$ by a factor of 2, which affects the Grid Mask and the patch-level loss $L_P$. We also investigate the effect of activating or deactivating the full image supervision $L_O$ on each possible combination.

We report the results of these experiments in Table~\ref{tab:ablation_p}.
Our proposed patch-level loss gives the additional supervision that is necessary for the network to stabilize and converge, mitigating the noise from the data interpolations. In addition, this patch-level supervision enables a form of visual representation learning, and the combination of it along with the image-level supervision yields the best performance. Furthermore, when evaluating the value of $P$, we found that a grid of $4\times4$ yield the best performance. 
We note that incrementing the value of $P$ hurts the performance dramatically. This may occur due to the significant level of freedom added by a $8\times8$ grid-mask, where the patch-level supervision is not able to mitigate the noise introduced. We also experimented on adding a hyperparameter to balance both losses but found out that giving the same weight performs better.

\begin{table}[th]
\begin{center}
\scalebox{0.75}{
\begin{tabular}{c|c|c|c}
\hline
 Grid & Image-level & Patch-level & Top-1 Acc \\
           & loss $L_O$ & loss $L_P$ &  \\
 \hline
 $2\times 2$ & \checkmark & \checkmark & 93.78 \\
 $2\times 2$ & \checkmark & \xmark & 93.30 \\
 $2\times 2$ & \xmark & \checkmark & 92.80 \\
 \hline
 $4\times 4$ & \checkmark & \checkmark & 94.10 \\
 $4\times 4$ & \checkmark & \xmark & 92.73 \\
 $4\times 4$ & \xmark & \checkmark & 92.02 \\
 \hline
 $8\times 8$ & \checkmark & \checkmark & 92.50 \\
 $8\times 8$ & \checkmark & \xmark & 91.94 \\
 $8\times 8$ & \xmark & \checkmark & 92.02 \\
 \hline
\end{tabular}
}
\end{center}
\caption{Ablation analysis: Top-1 accuracy on CIFAR-10 when varying the grid size, and the effect of using image and patch level supervision using ResNet-32 as the base network. }
\label{tab:ablation_p}
\end{table}

%% file: 053ablationFitnessFunc.tex
\subsection{Fitness Function Analysis.}
\label{sec:sup_additional_genetic_results}

We also evaluate how the fitness function impacts the performance of the combinations discovered by our genetic algorithm. After a network is trained using our Random PatchMix approach, it can be used as a function approximation of the underlying distribution generated by the image-pair interpolation along with their corresponding new labels. Thus, we can use this network to assess the patch mask $M_{i,j}$ configurations and class activations $(c_i,c_j)$, by computing the $L_P$ loss over these masks and image pairs.
We show our results in Table~\ref{tab:fitness_function}. 
We evaluate the how the patch-level accuracy of the validation set affects our genetic search. First, we show the result of 
using the configurations that yield the highest patch-level scores. Then we show the results of using the configurations that yield the lowest patch-level scores. We observe that using the configurations with the lowest patch-level accuracy yield better results. This means that the genetic algorithm is able to find challenging configurations for the model trained with random masks. Thus, our guided version allows the model to benefit from this information, leading to better results.

\begin{table}[h]
\setlength{\tabcolsep}{10pt}
\begin{center}
\scalebox{0.75}{
\begin{tabular}{c|c|c}
\hline
 Fitness  & \multicolumn{2}{c}{Allow Same Class Pairs?} \\
 Function & Yes & No      \\
 \hline
max $L_P$ & 94.80 & 95.42 \\
min $L_P$ & 95.97 & {\bf 96.32} \\
 \hline
\end{tabular}
}
\end{center}
\caption{Top-1 accuracy on CIFAR-10 when applying different fitness functions and the effect of using the same class combinations. We use PreAct-ResNet-164 as the backbone network architecture. }
\label{tab:fitness_function}
\end{table}

%% file: 070conclusion.tex
Our paper introduces PatchMix a novel interpolation method for augmenting the available number of samples during training by combining pairs of samples. Unlike previous methods that rely on patch-level interpolations our method allows for a more significant degree of flexibility regarding possible combinations by using a grid-like pattern. Moreover, an evolutionary search method for optimally selecting combinations that lead to increased exploration of critical areas of the input space was devised. We also found a fitness criteria that requires no model training by leveraging a pretrained PatchMix model that is trained by selecting random patches. We posit that PatchMix can serve as a regularizer that can complement other single sample data augmentation methods.

%% file: 080appendix.tex
\section{Supplementary Material}

\begin{figure*}[h]
\centering
  \includegraphics[width=0.90\textwidth]{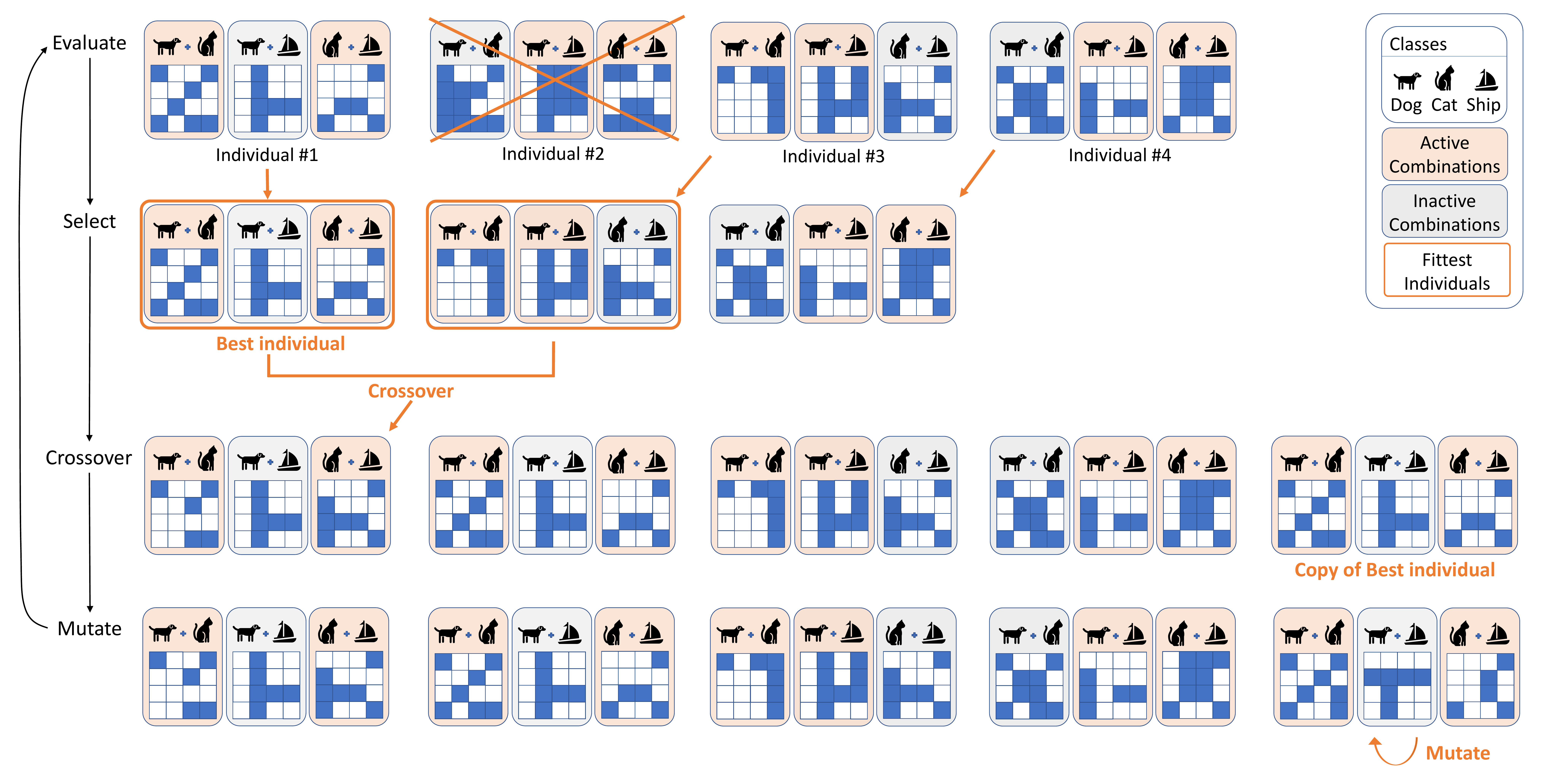}
\caption {Overview of our genetic search over patch combination strategies for a three-way dog, cat, ship classifier. First, we randomly sample a population of four individuals containing a set of Grid Mask configurations for all combinations. Then we evaluate them using our fitness function (i.e., our PatchMix algorithm), and the individuals with the less accurate predictions get selected. We randomly pick two individuals to crossover and create a new individual (i.e., offspring), and a copy of the best individual is mutated. The next cycle begins with the evaluation of the current pool of individuals, and the process repeats. }
\label{fig:fig_genetic_search}
\end{figure*}

\begin{figure}[h] 
\centering
  \includegraphics[width=0.48\textwidth]{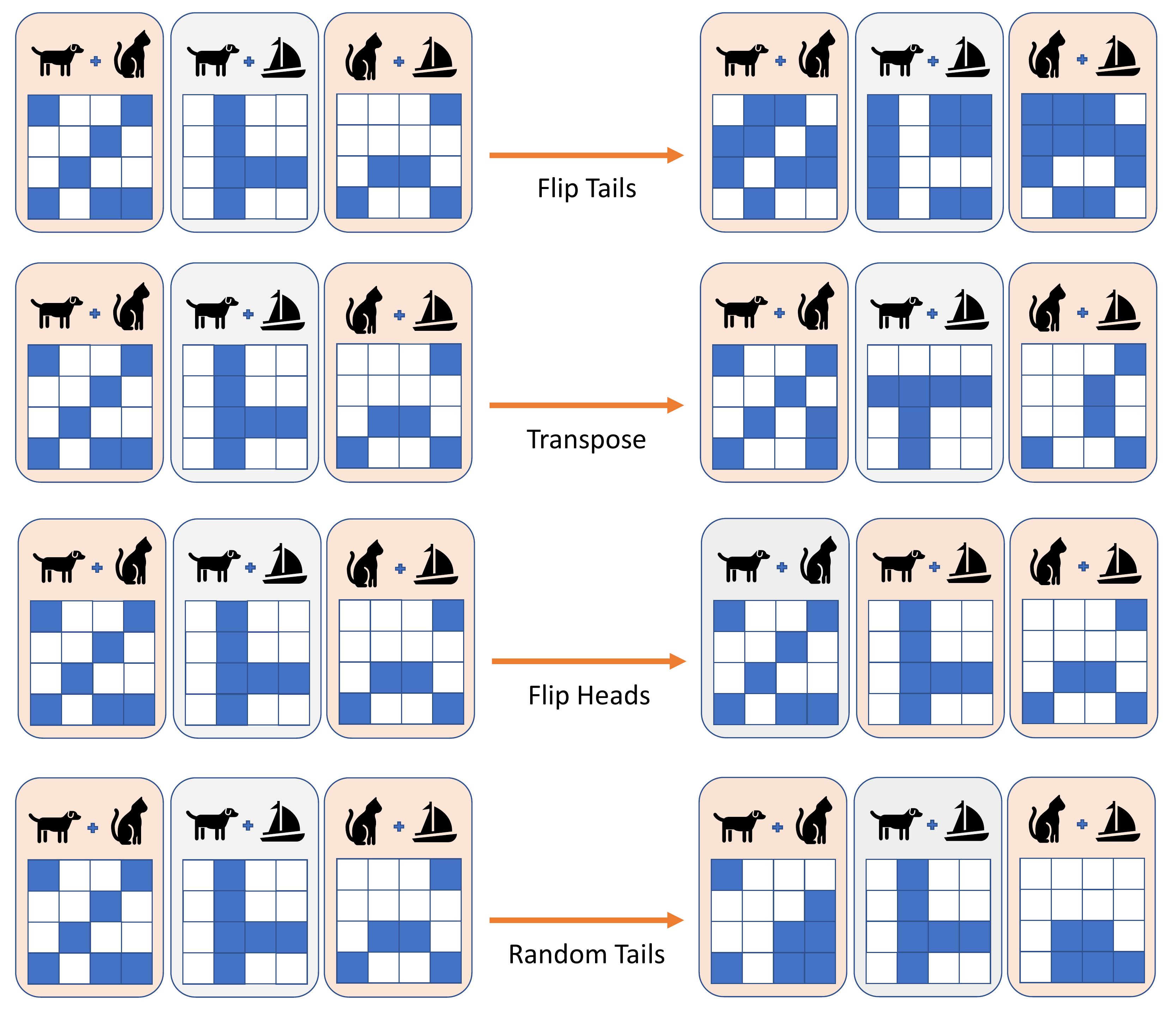}
\caption {Mutation examples. The original individual is in the left and its mutated version is in the right. The \textit{flip tails} operation exchange all the values from the active mask configurations. The \textit{transpose} operation switches the row and column indices of each patch. The \textit{flip heads} operation randomly exchange the active configurations. The \textit{random tails} operation randomly activate and deactivate some values of the active masks. }
\label{fig:fig_mutations}
\vspace{0.2in}
\end{figure}

First, we illustrate an example of our genetic search approach for three classes in Figure~\ref{fig:fig_genetic_search}, and we also show some mutation examples in Figure~\ref{fig:fig_mutations}. In section~\ref{sec:sup_manifold_intrusion} we include further analysis for the manifold intrusion problem. Next, we show some qualitative examples for the Weakly Supervised Object Localization and Object Detection experiment using the Class Activation Mapping (CAM) in \ref{sec:qualitative_WSOL}. Additional ablation studies for PatchMix varying architectures, the supervision effect of the patches, and the grid size (number of $P$ patches) are shown in section~\ref{sec:ablation_studies}. We also include results when varying the genetic search fitness function and training criterion for our guided PatchMix approach in section~\ref{sec:sup_additional_genetic_results}. We show in-depth results of the search performance of our genetic algorithm for CIFAR-10 in section~\ref{sec:add_search_performance}. Finally, we discuss the computational overhead of PatchMix compared to other similar interpolation approaches in section~\ref{sec:computational_overhead}.

\subsection{Manifold Intrusion Analysis.}
\label{sec:sup_manifold_intrusion}

\begin{figure*}[h]
\begin{center}
  \includegraphics[width=.90\textwidth]{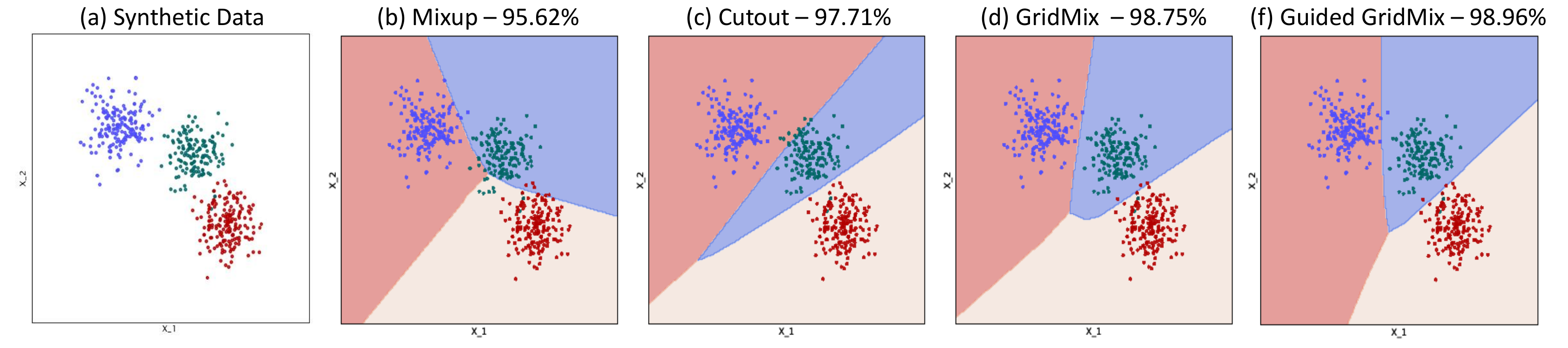}
\caption{ Effect of different data interpolation techniques on the decision boundary for a classification task. 
}
\label{fig:add_decision_boundaries}
\end{center}
\end{figure*}

We show in Figure~\ref{fig:add_decision_boundaries} the effect in the decision boundary for a three-way classifier on synthetic data when using different interpolations such as Mixup, Cutout, and our proposed PatchMix.

The main advantages of PatchMix include preventing the over-sampling of synthetic data that is prone to suffer the \textit{manifold intrusion} problem during training.
We train a two-layer neural network to classify a toy dataset with three linearly separable classes. Since we are using only two features per class, the Cutmix and Random PatchMix interpolations behave similarly (for both cases, the random masks have only two choices: $0\vert0$, $1\vert0$, $0\vert1$ or $1\vert1$). While all interpolation techniques smooth the decision boundaries, Mixup and Cutout suffer from the \textit{manifold intrusion} problem, hurting the model. Our guided version of PatchMix induces a better decision boundary by adding predefined combinations based on difficult mixed samples.

\subsection{Weakly Supervised Object Localization and Object Detection.}
\begin{figure*}[h] 
\centering
  \includegraphics[width=.9\textwidth]{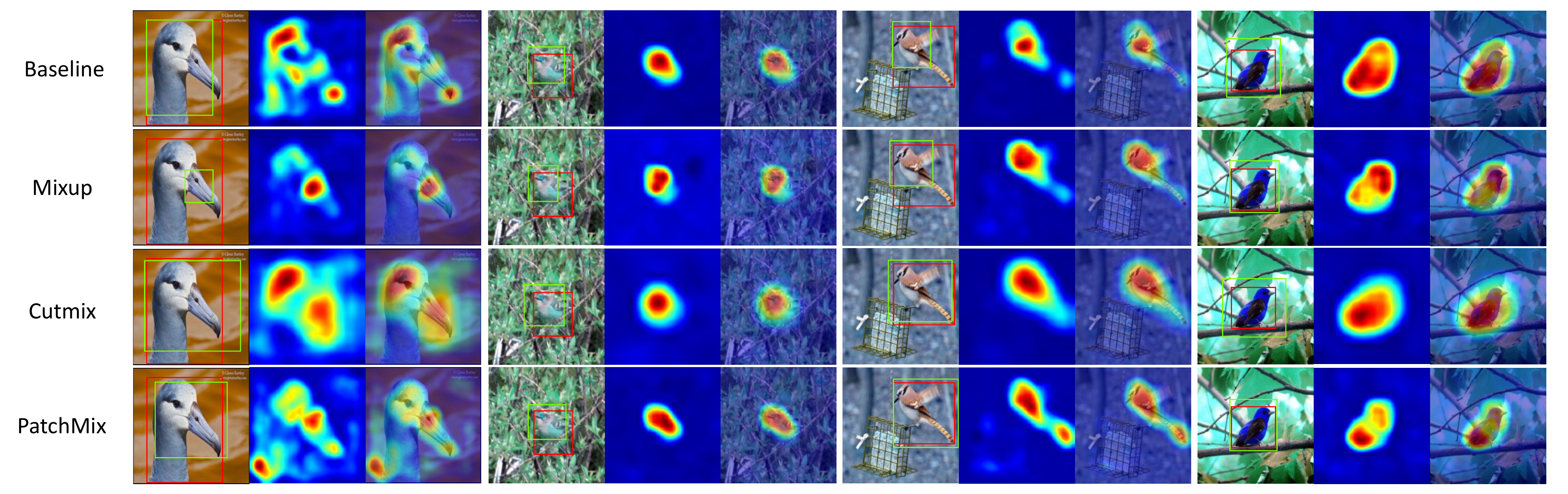}
\caption {Weakly supervised object localization task on CUB-200-2011 dataset trained on ResNet-50. The ground truth bounding boxes are shown in red and the predicted bounding boxes are shown in light green.}
\label{fig:fig_wsol}
\end{figure*}

Figure \ref{fig:fig_wsol} shows some results for the weakly supervised object localization task on CUB-200-2011 dataset trained on ResNet-50. We show qualitative results when using the CAM baseline, Mixup, CutMix and PatchMix. The ground truth bounding boxes are shown in red and the predicted bounding boxes are shown in light green. We use the Class Activation Mapping (CAM) to extract the attention maps, and then we compute the maximal box accuracy, which is the bounding box accuracy and the Intersection over Union (IoU) of the proposed boxes.
\label{sec:qualitative_WSOL}

%% file: 051searchAnalysis.tex
\subsection{Analysis on Search for Guided Pairs \& Masks} \label{sec:search_performance}

\begin{figure}[!th] 
\centering
  \includegraphics[width=0.4\textwidth]{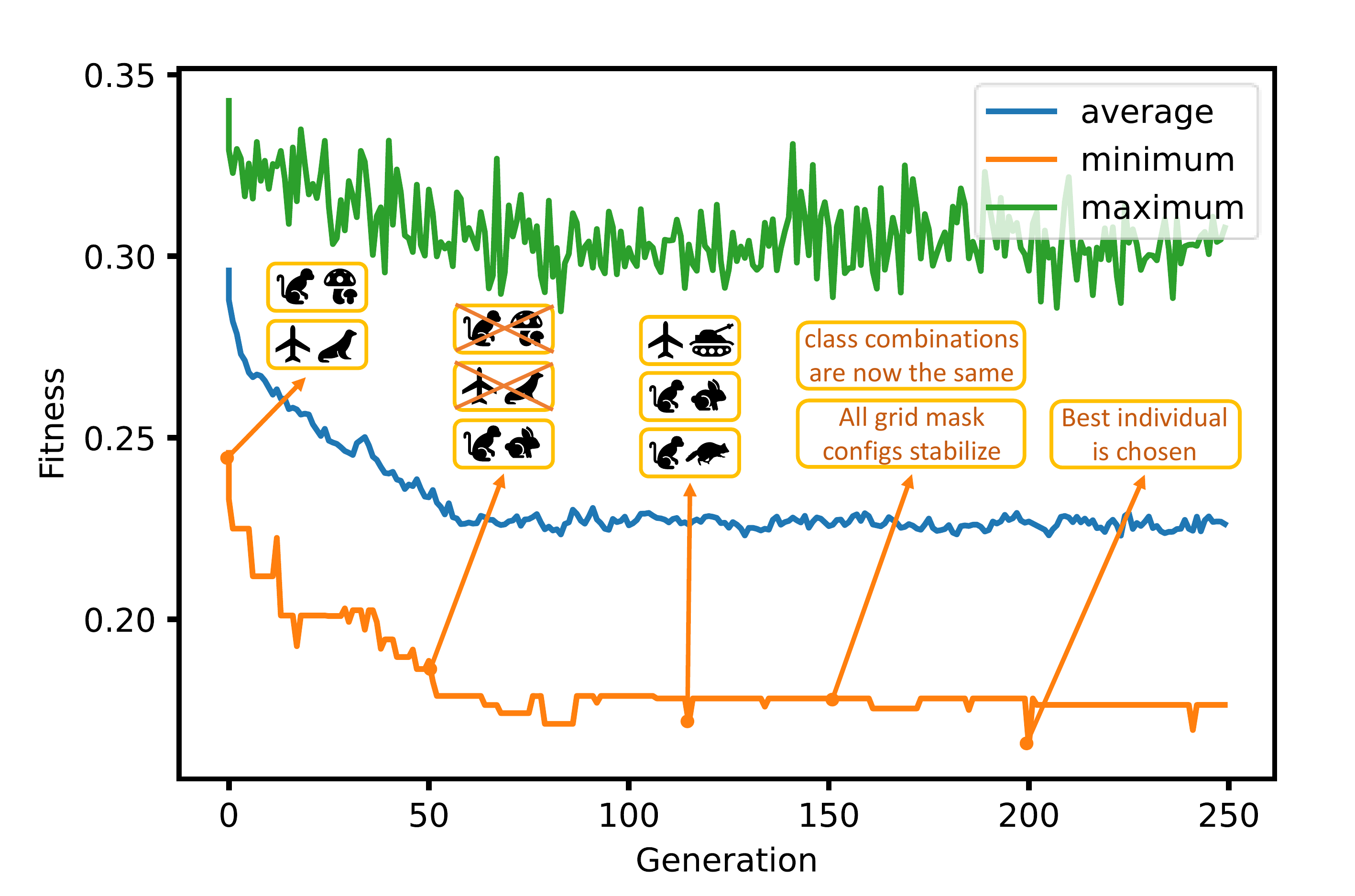}
\caption { Performance of our genetic search after 250 generations on CIFAR-100. The \textit{y} axis shows the top-1 accuracy of the population evaluated on our fitness function.
We highlight some class combinations included and excluded every 50 generations. }
\label{fig:fig_search_performance}
\vspace{-0.12in}
\end{figure}

Figure~\ref{fig:fig_search_performance} shows the evolution of our fitness function when searching for the grid mask configurations on CIFAR-100 over $250$ generations. Since we are looking for the most informative samples, we want to find the class combinations and grid configurations for which a pretrained Random PatchMix struggles the most. Hence, our criteria for fitness allows the discovery of individuals that contribute with more information as explained in our method section. We observe that during the search process, some class combinations get discarded systematically in the search process. For instance, the model initially selects the pairs (\emph{plain}, \emph{seal}), (\emph{chimpanzee}, \emph{mushroom}), but those are discarded after 50 generations and more informative combinations 
are selected. 
After 100 generations, the model has discovered many of the class combinations that it will use, such as (\emph{chimpanzee}, \emph{raccoon}), (\emph{road}, \emph{tractor}), and (\emph{sea}, \emph{shark}). 

Evolutionary search has been used in previous work for neural architecture search (NAS) where a space of possible network designs is explored, and can be extended for data augmentation where a combinatorial space of image transformations are explored. 
In both cases, the bottleneck is to evaluate each configuration -- as it requires training a deep neural network to assess these choices. Our work goes beyond that as we are able to entirely bypass any amount of training. Our idea of using a PatchMix model pretrained on random configurations to define a fitness criteria allows for this to happen.

%% file: 081appendix.tex
\subsection{Additional Ablation Studies.}
\label{sec:sup_additional_aablation}

In this section we show additional results when varying the grid size (number of $P$ patches), the model architecture and whether the image or patch supervision are used in the training process. Table~\ref{tab:additional_ablation_p} shows that using the image level supervision along with the patch level supervision consistently outperforms all other variations. Furthermore, in all architectures a grid size of $4\times 4$ outperforms other grid size possible selections.

\begin{table}[th]
\begin{center}
\begin{tabular}{c|c|c|c|c}
\hline
 & Grid & Image-level & Patch-level & Top-1 Acc \\
 &           & loss $L_O$ & loss $L_P$ &  \\
 \hline
 \multirow{9}{*}{\STAB{\rotatebox[origin=c]{90}{PreActResNet-56}}}
 & $2\times 2$ & \checkmark & \checkmark & {\bf 94.53} \\
 & $2\times 2$ & \checkmark & \xmark & 93.78 \\
 & $2\times 2$ & \xmark & \checkmark & 94.03 \\
 \cline{2-5}
 & $4\times 4$ & \checkmark & \checkmark & {\bf 95.28} \\
 & $4\times 4$ & \checkmark & \xmark & 94.97 \\
 & $4\times 4$ & \xmark & \checkmark & 94.73 \\
 \cline{2-5}
 & $8\times 8$ & \checkmark & \checkmark & {\bf 94.08} \\
 & $8\times 8$ & \checkmark & \xmark & 92.84 \\
 & $8\times 8$ & \xmark & \checkmark & 92.57 \\
 \midrule
 \multirow{9}{*}{\STAB{\rotatebox[origin=c]{90}{PreActResNet-164}}}
 & $2\times 2$ & \checkmark & \checkmark & {\bf 94.55} \\
 & $2\times 2$ & \checkmark & \xmark & 94.27 \\
 & $2\times 2$ & \xmark & \checkmark & 94.53 \\
 \cline{2-5}
 & $4\times 4$ & \checkmark & \checkmark & {\bf 96.32} \\
 & $4\times 4$ & \checkmark & \xmark & 94.56 \\
 & $4\times 4$ & \xmark & \checkmark & 94.37 \\
 \cline{2-5}
 & $8\times 8$ & \checkmark & \checkmark & {\bf 94.60} \\
 & $8\times 8$ & \checkmark & \xmark & 94.03 \\
 & $8\times 8$ & \xmark & \checkmark & 93.86 \\
 \hline
\end{tabular}
\end{center}
\caption{Ablation analysis: Top-1 accuracy on CIFAR-10 when varying the grid size, and the effect of using image and patch level supervision using different network backbones. }
\vspace{0.1in}
\label{tab:additional_ablation_p}
\end{table}

\subsection{Additional Search Performance} \label{sec:add_search_performance}

\begin{figure}[t] 
\begin{center}
  \includegraphics[width=0.45\textwidth]{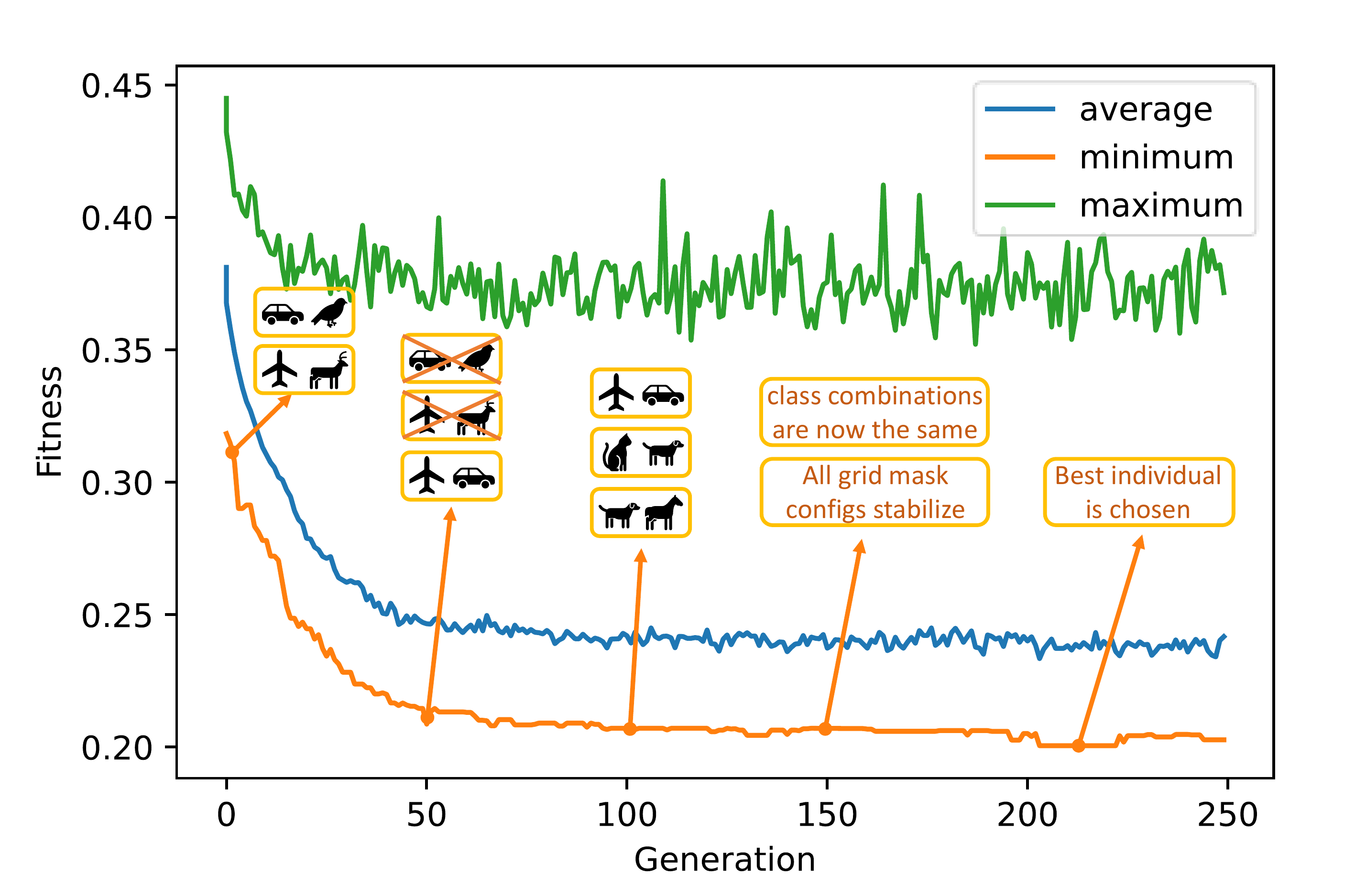}
\caption { Performance of our genetic search after 250 generations on CIFAR-10. The \textit{y} axis shows the top-1 accuracy of the population evaluated on our fitness function, which is a network trained with PatchMix and random sampling. For this example, the architecture of the network is PreAct-ResNet-164. We highlight some class combinations included and excluded every 50 generations. }
\label{fig:fig_search_performance2}
\end{center}
\end{figure}

Figure~\ref{fig:fig_search_performance2} shows how the evolution of our fitness function when searching for the grid mask configurations on CIFAR-10 over $250$ generations. Since we are looking for the most difficult samples, we want to find the class combinations and grid configurations for which a pretrained model using PatchMix with random sampling struggles the most. Hence our criteria for fitness is individuals that score the lowest under this criteria as explained in our method section. We observe that when searching the space, even the score of the individuals with the best top-1 accuracy drops, which is expected in our setup. We also observe that during the search process, some class combinations get discarded systematically in the search process. For instance, the model initially selects the pair (\emph{automobile}, \emph{bird}), and (\emph{airplaine}, \emph{deer}), but those pairs are discarded after 50 generations and more informative combinations such as (\emph{automobile}, \emph{airplane}) are selected. After 100 generations, the model has already discovered many of the class combinations that it will use such as (\emph{automobile}, \emph{airplane}), (\emph{dog}, \emph{horse}), (\emph{dog}, \emph{cat}), and after 150 epochs converges to a set of categories and the grid mask patterns stabilize as well. Finally, at around epoch 200, the best individual containing the best pairs and mask patterns is chosen.

\subsection{Impact of Selected Number of Configurations}
We also show in Table~\ref{tab:number_configurations} the impact of using different amounts of active combinations when evaluating our approach with CIFAR-100. In general, the search algorithm performs better when setting a limited amount of active combinations $N$ equal to the number of classes. We also observe that the best results are achieved when using only the discovered class combinations. This also supports our rationale about the positive gains of the model when training using the configurations discovered by our genetic algorithm. In general, our evolutionary formulation is effective and yields the best results.

\subsection{Computational Overhead}
\label{sec:computational_overhead}
Random PatchMix does not add a significant overhead -our mask is randomly sampled-, and the additional patch-level branches make the model $\approx20\%$ larger; i.e., a modified model did not surpass any of the available GPU capacity during our experiments. 
For our guided version, the overhead depends on the number of classes and active combinations; i.e., for CIFAR10, using a set of 10 parallel processes to assess the fitness criterion, it would take $\approx4$ hours to find the optimum combination. This can be further decreased for at least $\approx75\%$ by using a subset of the val.~set randomly sampled from the whole val.~set, shrinking the initial population, and reducing the number of generations (we found that the genetic algorithm stabilizes around the 60th generation, but we continue searching for 250 generations). During test time, Guided-PatchMix, does not add any significant overhead compared to other similar methods.

\begin{table}[h]
\setlength{\tabcolsep}{10pt}
\begin{center}
\begin{tabular}{c|c|c}
\hline
 $N$ Comb  & \multicolumn{2}{c}{Allow Same Class Pairs?} \\
           & Yes & No      \\
 \hline
200 & 77.00 & {\bf 78.53} \\
300 & 76.61 & 77.55 \\
1000 & 64.98 & 65.54 \\
 \hline
\end{tabular}
\end{center}
\caption{Top-1 accuracy on CIFAR-100 when varying the number of active class combinations. We also show the effect of using the same class pairs. We use PreAct-ResNet-164 as the backbone network architecture. }
\label{tab:number_configurations}
\end{table}

\begin{figure}[t!] 
\begin{center}
  \includegraphics[width=0.35\textwidth]{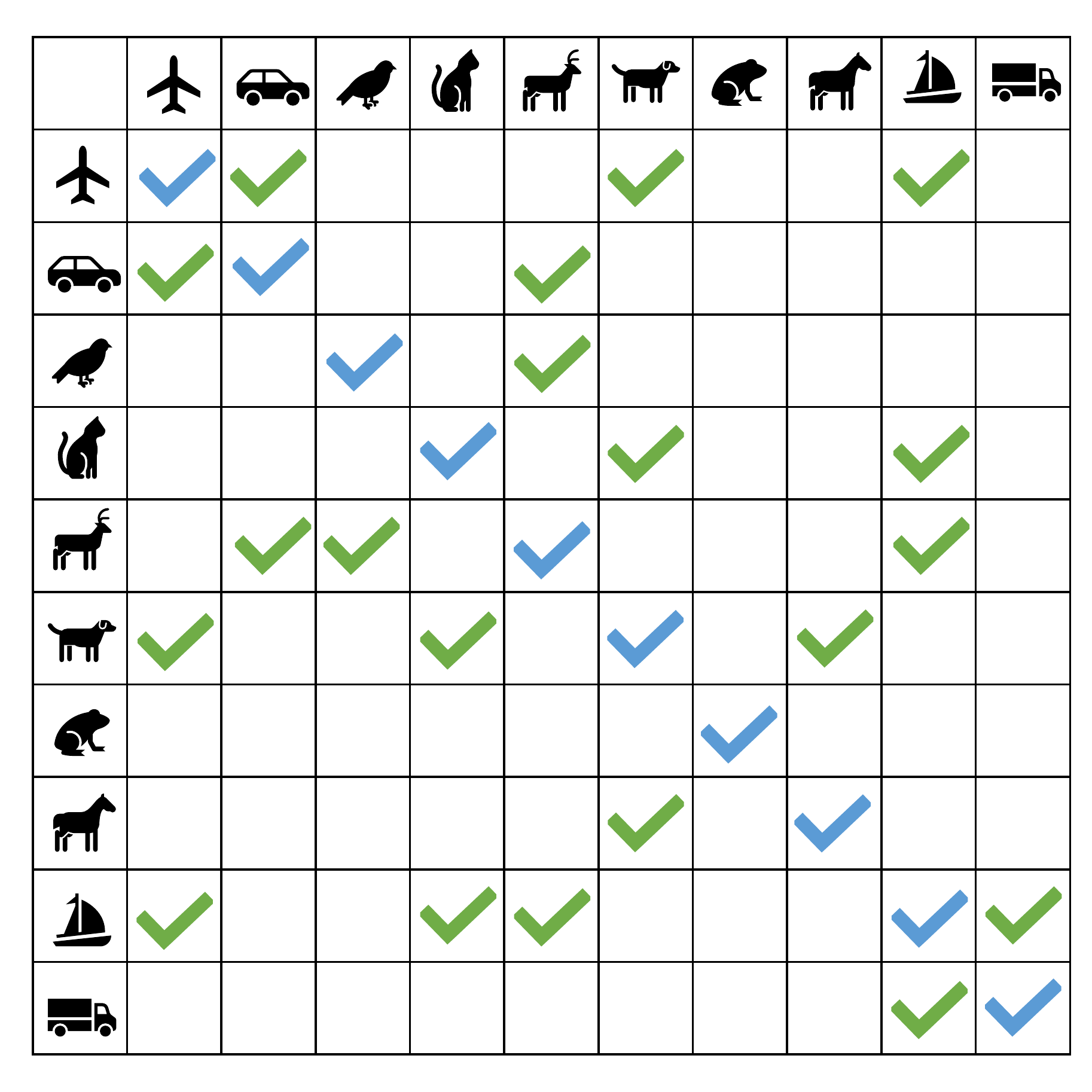}
\caption { Class combinations. The green checks correspond to the class combinations found by our search algorithm. The blue checks correspond to the classes we force to always appear in the combinations. 
}
\label{fig:combs}
\end{center}
\end{figure}

Figure~\ref{fig:combs} shows the final top 10 class combinations found by the genetic algorithm on CIFAR-10. When training using the guided approach, the combination of these classes are more likely to be sampled. The evolutionary process selects many pairs that are semantically close such as (\emph{dog}, \emph{cat}), (\emph{dog}, \emph{horse}), (\emph{ship}, \emph{truck}), or (\emph{ship}, \emph{airplane}). 

\subsection{Additional Implementation Details}
In this section we present the pseudo-code to train our random and guided steps four our PatchMix approach. We first train a network using 
Algorithm~\ref{alg:RandomPM} in which we divide the last convolutional layer of our network to output the same amount of patches like the ones defined in the input. For example, the convolutional layer of the latest block in a ResNet50 will output a tensor of shape $[N, C_{out}, H_{out}, W_{out}]$; if the input corresponds to images of $224x224x3$ and we divide our image in 16 patches $(P=4)$, $C_{out}=2048$ denotes the number of channels and $H_{out}=28$, $W_{out}=28$ are the height and width of this output. Thus we can also divide this output into a grid-like layer in which each patch would be of size $7x7$. We then apply the average pooling over each of these patches and finally the linear transformation to output the probability distribution of each patch prediction. This network also outputs a separate branch that takes into account the whole convolutional layer to predict the probability distribution of each input image as a whole. We then use genetic search, which we define in Algorithm~\ref{alg:GeneticSearch} to find the best masks $M_{i,j}$ and category pairs $(c_i, c_j)$ that correspond to each of the discovered class combinations. We then use $M_{i,j}$ to augment the training samples based on the class combinations $(c_i, c_j)$ discovered by our genetic search.

\begin{algorithm}[h!] 
\caption{Pseudo-code of our first step: Random PatchMix}\label{alg:RandomPM}
\begin{algorithmic}[1]
\State \textbf{Require:} {$P$}

\For{(x, y) in (data\_loader)}
\State index = torch.randperm(batch\_size)
\State {lam = random.beta(1, 1)}
\State {quad\_mask\_vals = random.beta(1, 1, size=[P, P]).round()}
\State {quadrant\_size = int(image\_size/P)}
\State {result = []}
\For {r in range(P)}
\State {    rows = full((quadrant\_size,quadrant\_size), quad\_mask\_vals[r,0])}
    \For {c in range(1, P)}
\State {        row\_column = full((quadrant\_size,quadrant\_size), quad\_mask\_vals[r,c])}
\State {        rows = concatenate((rows, row\_column), axis=1)}
    \EndFor
\State {    result.append(rows)}
\EndFor
\State {image\_mask = result.reshape(image\_size,image\_size)}
\State {layer\_mask = result.reshape(layer\_size,layer\_size)}
\State {mixed\_x = zeros((x.shape))}
\State {mixed\_x = x*image\_mask + x[index]*(1-image\_mask)}
\State {new\_y = []}
\State {layer\_mask = list(layer\_mask.flatten())}
\For {m in layer\_mask}
    \If {m == 0}
    \State {new\_y.append(y[index])}
    \EndIf
    \State {\textbf{else}: new\_y.append(y)}
\EndFor
\State {lam = layer\_mask.count(1)/len(layer\_mask)}
\State {patch\_output, image\_output = model(mixed\_x)}
\State {$L_P$ = patch\_level\_criterion(patch\_output, new\_y, class\_criterion, grid\_range = $P^2$)}
\State {$L_O$ = image\_level\_criterion(class\_criterion, image\_output, y, y[index], lam)}
\State {loss = ($L_O$ + $L_P$) / 2}
\State {update(model)}
\EndFor
\State \textbf{end}
\end{algorithmic}
\end{algorithm}

\begin{algorithm}[h!] 
\caption{Pseudo-code of our Evolutionary Search}\label{alg:GeneticSearch}
\begin{algorithmic}[1]
\State {head = get\_array\_activations()}
\State {tail = get\_array\_configurations()}
\State {population = []}
\For {i in range(total\_pop)}
\State {        tmp\_indiv = zeros(head)}
\State {        selected\_head = random.sample(range(index\_head), new\_comb\_count)}
\State {        tmp\_indiv[selected\_head] = 1}
\State {        tmp\_indiv[default\_head] = 1 }
\State {        tmp\_indiv\_tail = random.randint(0, 2, tail)}
\State {        individual = list(concatenate((tmp\_indiv, tmp\_indiv\_tail)))}
\State {        individual = creator(individual)}
\State {        population.append(individual)}
\EndFor
\State {toolbox = DEAP.Toolbox()}
\State {toolbox.register("population", load\_population(), population\_size)}
\State {toolbox.register("evaluate", eval\_population())}
\State {toolbox.register("mate", crossover(), prob=0.50)}
\State {toolbox.register("mutate", mut\_ops(), prob=0.30)}
\State {toolbox.register("select", toolbox.selTournament, tournsize=3)}
\For {gen in range(0, args.generations)}
\State {        offspring = toolbox.select(population, len(population))}
\State {        offspring = algorithms.varAnd(offspring, toolbox, cxpb, mutpb)}
\State {        invalid\_ind = [ind for ind in offspring if not ind.fitness.valid]}
\State {        fitnesses = toolbox.map(toolbox.evaluate, invalid\_ind)}
        \For {ind, fit in zip(invalid\_ind, fitnesses)}
        \State {    ind.fitness.values = fit}
        \EndFor
        
        \If{halloffame is not None}
         \State {   halloffame.update(offspring)}
        \EndIf
\State {        population[:] = offspring}
\EndFor

\State \textbf{end}
\end{algorithmic}
\end{algorithm}